\documentclass[11pt]{article}

\usepackage[preprint]{acl}

\makeatletter
\acl@anonymizefalse
\makeatother

\usepackage{times}
\usepackage{latexsym}

\usepackage[T1]{fontenc}

\usepackage[utf8]{inputenc}

\usepackage{microtype}

\usepackage{inconsolata}

\usepackage{graphicx}

\usepackage{booktabs}
\usepackage{multirow}
\usepackage[normalem]{ulem}
\useunder{\uline}{\ul}{}
\usepackage{booktabs}
\usepackage{enumitem}
\usepackage{xcolor}
\usepackage{tikz}
\usepackage[most]{tcolorbox}
\usepackage{graphicx}
\usepackage{pgfplots}
\pgfplotsset{compat=1.18}
\usepgfplotslibrary{groupplots}
\usepackage{filecontents}
\usepackage[table,xcdraw]{xcolor}

\usepackage{amsthm}
\usepackage{amssymb}
\usepackage{subcaption}
\usepackage{colortbl}
\usepackage{pifont}
\usepackage{wrapfig}
\usepackage[table]{xcolor}
\usepackage{booktabs}
\usepackage{array}
\usepackage{pifont}
\usepackage{makecell}
\usepackage{amsmath}
\usepackage{multirow}
\usepackage{algorithm}
\usepackage{algorithmic}
\usepackage{listings}

\usepackage[most]{tcolorbox}
\usepackage{array}
\usepackage{booktabs}
\usepackage[table]{xcolor}

\lstdefinestyle{promptstyle}{
    frame=tb,
    basicstyle=\ttfamily\footnotesize,
    columns=fullflexible,
    breaklines=true,
    postbreak=\mbox{\textcolor{red}{$\hookrightarrow$}\space},
    captionpos=b,
    showstringspaces=false,
    numbers=none,
}

\usepackage{float}
\usepackage{siunitx}
\usepackage{enumitem}
\setlist[itemize]{leftmargin=*, noitemsep, topsep=0pt}

\AtBeginDocument{%
  \setlength{\abovedisplayskip}{4pt plus 1pt minus 1pt}
  \setlength{\belowdisplayskip}{4pt plus 1pt minus 1pt}%
  \setlength{\abovedisplayshortskip}{2pt plus 1pt}%
  \setlength{\belowdisplayshortskip}{2pt plus 1pt}%
  \setlength{\bibsep}{3pt}
}

\definecolor{topone}{RGB}{224,210,255}    
\definecolor{toptwo}{RGB}{237,226,255}    
\definecolor{topthree}{RGB}{247,240,255}  

\newcommand{\best}[1]{\cellcolor{topone}\textbf{#1}}
\newcommand{\second}[1]{\cellcolor{toptwo}#1}
\newcommand{\third}[1]{\cellcolor{topthree}#1}

\definecolor{sirinbg}{RGB}{250,247,255}      
\definecolor{sirinhl}{RGB}{232,222,252}      
\definecolor{sirinrule}{RGB}{165,132,210}    
\definecolor{sirintext}{RGB}{38,32,48}       
\definecolor{sirinmint}{RGB}{222,248,240}    
\definecolor{sirinpink}{RGB}{252,225,238}    

\definecolor{headerbg}{HTML}{EAF1F8}
\definecolor{ourbg}{HTML}{E8F5E9}
\definecolor{yesbg}{HTML}{D9EAD3}
\definecolor{partbg}{HTML}{FFF2CC}
\definecolor{nobg}{HTML}{F4CCCC}
\definecolor{lightrow}{HTML}{F8F9FA}

\newcommand{\cmark}{\ding{51}}
\newcommand{\xmark}{\ding{55}}

\newcommand{\capyes}{\cellcolor{topone}{\color{sirintext}\cmark}}
\newcommand{\cappart}{\cellcolor{topthree}{\color{sirinrule}$\sim$}}
\newcommand{\capno}{{\color{black!25}\xmark}}

\hypersetup{colorlinks=true, linkcolor=blue!50!black, citecolor=blue!50!black, urlcolor=blue!50!black}

\lstdefinestyle{pyc}{
  basicstyle=\ttfamily\footnotesize,
  language=Python,
  showstringspaces=false,
  breaklines=true,
  keywordstyle=\color{blue!70!black}\bfseries,
  commentstyle=\color{green!45!black}\itshape,
  stringstyle=\color{red!60!black},
  numbers=none,
  frame=tb,
  framesep=3pt,
  xleftmargin=2pt,
  xrightmargin=2pt
}

\newcommand{\framework}{\textsc{SIRIN}}

\newcommand{\sirinlogo}{%
  \raisebox{-0.4em}{\includegraphics[height=1.8em]{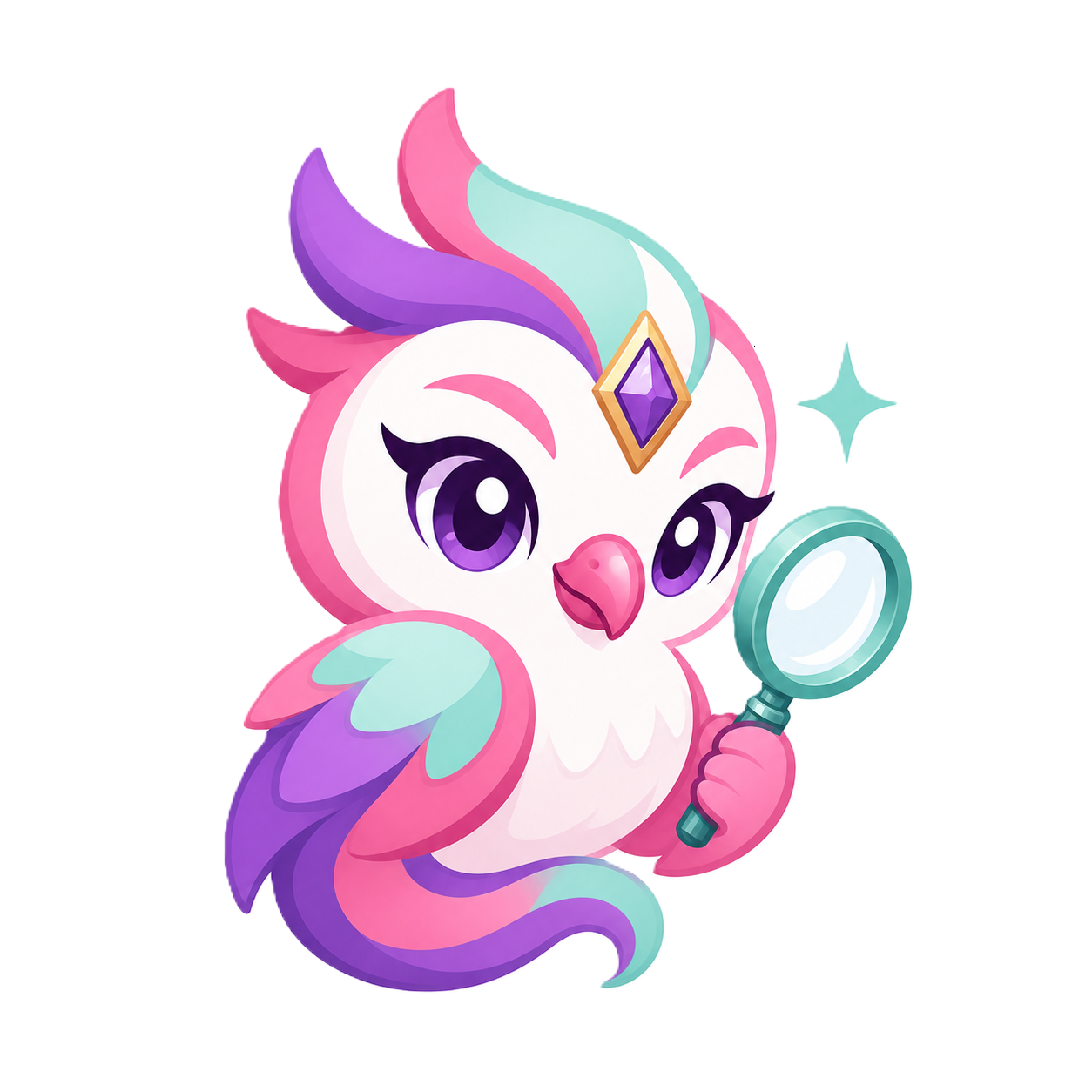}}%
}

\title{\sirinlogo\ \textbf{\framework{}}: A Unified Toolkit for Detecting Contextual Hallucinations in Retrieval-Augmented and Memory-Grounded LLM Systems}

\author{
  \textbf{Julia Belikova \quad Rauf Parchiev \quad Mikhail Filimonov \quad Konstantin Polev} \\
  \textbf{Andrey Savchenko \quad Maksim Makarenko} \\
  Sber AI Lab \\
}


\begin{document}
\maketitle

\begin{abstract}
\framework{} (Semantic Inconsistency Recognition and Inspection Nexus) is a unified toolkit and interactive web UI for detecting contextual hallucinations (fluent, plausible responses unsupported by the provided evidence) in retrieval-augmented, agentic, and memory-grounded LLM systems. \framework{} unifies three detector paradigms (representation probing, uncertainty estimation, and judge-style verification) and the complementary task of pre-generation query answerability under one interface, configuration system, and evaluation pipeline, supporting response- and span-level inspection in both white-box and black-box settings. The web UI enables live analysis of user-supplied context--query--answer triples through hallucination scores, unsupported-span highlighting, and side-by-side detector comparison, with a lightweight plug-in design for adding new detectors. We demonstrate \framework{} on hallucination detection, query answerability, and as a faithfulness gate within long-term memory systems. 

The source code is publicly available at \url{https://github.com/sb-ai-lab/SIRIN}.
\end{abstract}

\section{Introduction}
\label{sec:intro}

Large language models (LLMs) often produce answers that are fluent and plausible, yet not grounded in the evidence they are given~\cite{ji2023survey,huang2025survey}. In context-grounded generation, this failure is usually framed as \emph{contextual hallucination}: the model introduces claims that are unsupported by, or inconsistent with, the provided context~\cite{ragtruth,lookback2024}. Such failures are especially consequential in deployed RAG systems and in agentic pipelines, where models repeatedly retrieve, summarize, store, and update declarative or procedural memory over long-term interaction~\cite{agenthallucinations2025,simplemem2026}.

\begin{table}[t]
\centering
\small
\setlength{\tabcolsep}{3.4pt}
\renewcommand{\arraystretch}{1.18}
{\arrayrulecolor{sirinrule}
\begin{tabular}{@{}lcccccc@{}}
\toprule
\textbf{Toolkit} & \textbf{Probe} & \textbf{Judge} & \textbf{UE} &
\textbf{Ans.} & \textbf{Span} & \textbf{UI} \\
\midrule

LM-Polygraph
& \cappart & \capno & \capyes & \capno & \cappart & \cappart \\

\rowcolor{sirinbg}
\cellcolor{white}RAGAS
& \capno & \capyes & \capno & \cappart & \cappart & \capno \\

DeepEval
& \capno & \capyes & \capno & \cappart & \cappart & \capno \\

\rowcolor{sirinbg}
\cellcolor{white}UpTrain$^\dagger$
& \capno & \capyes & \capno & \capyes & \cappart & \capyes \\

TruLens
& \capno & \capyes & \capno & \cappart & \cappart & \capyes \\

LettuceDetect
& \capno & \capyes & \capno & \capno & \capyes & \capyes \\

\midrule

\textbf{\framework{}} (this work)
& \capyes & \capyes & \capyes & \capyes & \capyes & \capyes \\

\bottomrule
\end{tabular}
}
\caption{Comparison with representative open-source evaluation toolkits.
Probe = trained probes over model internals; UE = uncertainty estimation;
Ans.\ = query answerability; Span = sub-response localization.
\cmark{} = native support; $\sim$ = partial support, such as claim- or
statement-level verdicts rather than span offsets, query--context relevance
rather than an explicit answerability decision, density estimators over hidden
states rather than trainable probes, or a deprecated demo UI; \xmark{} = absent.
$^\dagger$No public development activity in the past two years.}
\label{tab:toolkit_comparison}
\end{table}

This need has led to a broad and heterogeneous landscape of hallucination detectors, which largely fall into three families. \textbf{Probing} methods train lightweight classifiers over generator-side signals such as hidden states, attention maps, and token probabilities, from representation probes to attention-based methods such as Lookback Lens~\cite{ji2024internal,lookback2024}. \textbf{Judge}-style methods rely only on observable inputs and outputs, using sampled generations, semantic consistency checks, encoder-based classifiers, or LLM-as-a-judge verification~\cite{selfcheck2023,ragas,lettucedetect2025,faithjudge2025}. \textbf{Uncertainty estimation} (UE) scores generations by the model's own confidence, from likelihood-based scores such as perplexity to sampling-based semantic uncertainty and attention-aware estimators such as RAUQ~\cite{fadeeva2023,semanticentropy2024,vazhentsevefficient}. Two axes cut across the families: the model access they assume (white-box internals versus black-box text) and the granularity at which they operate, where response-level scores support runtime gating while entity- and span-level predictions are needed for inspection, debugging, and user-facing explanations~\cite{ragtruth,yeh2025halluentity,psiloqa2025,mushroom2025}.

Despite this progress, using these methods in a single LLM system remains cumbersome: detector implementations differ in data format, configuration style, backend assumptions, supported granularity, and evaluation protocol, which obscures comparison and complicates combining complementary signals across pipeline stages. Such signals span the whole generation cycle, from pre-generation checks of whether the context is sufficient to answer a query (\emph{query answerability}) to post-generation checks of whether the produced answer is faithful to that context. Existing open-source toolkits reflect this fragmentation rather than resolving it (Table~\ref{tab:toolkit_comparison}). LM-Polygraph~\cite{fadeeva2023,lmpolygraph2025} offers the most comprehensive uncertainty-estimation catalog but provides neither trainable probes nor judge-based verification. 
RAGAS~\cite{ragas}, DeepEval~\cite{deepeval}, UpTrain~\cite{uptrain}, and TruLens~\cite{trulens} primarily rely on LLM-as-a-judge verification, exposing at most statement-level verdicts, with no access to generator internals or uncertainty signals. LettuceDetect~\cite{lettucedetect2025} instead uses a fine-tuned encoder to predict token-level hallucinated spans, giving finer localization but still confined to a single detector regime. To date, no existing toolkit unifies all three families behind a single interface that covers both answerability and faithfulness at response and span granularity, together with an interactive inspection UI.

We introduce \textbf{\framework{}} (\textbf{S}emantic \textbf{I}nconsistency \textbf{R}ecognition and \textbf{I}nspection \textbf{N}exus), a toolkit that addresses this fragmentation by treating contextual hallucination detection as an infrastructure problem rather than a single-detector problem. \framework{} provides a common environment for developing new detectors, inspecting and benchmarking existing ones, and deploying them in practical LLM systems, from query answerability before generation to answer faithfulness verification after generation. The accompanying web UI\footnote{Demo video: \url{https://rebrand.ly/SIRIN}} makes this concrete: a visitor pastes a context--query--answer triple, selects detectors from any of the three families, and sees a calibrated hallucination score, unsupported spans highlighted on the answer, and a side-by-side comparison of detector outputs (Figure~\ref{fig:ui-main}).

The main contributions of this work are:
\begin{itemize}
    \item \framework{}, a \textbf{unified open-source toolkit} that brings probing-based methods, judge-style verification, and uncertainty estimation, together with the complementary task of query answerability, under a shared interface, data format, configuration system, and evaluation pipeline, covering white-box and black-box regimes at response and span granularity with plug-in extension points for new detectors. The same detectors could be attached as a drop-in trust layer around the retrieval stage of agentic memory modules, gating answers on both answerability and faithfulness (\S\ref{sec:simplemem}).

    \item An \textbf{interactive web interface} for practical hallucination inspection, including live analysis of context--query--answer triples, hallucination scores, unsupported-span visualization, and side-by-side comparison of detector outputs, available at \url{https://hf.co/spaces/parchiev/SIRIN}.

    \item An \textbf{empirical demonstration} of \framework{} on contextual hallucination detection and query answerability, together with its integration as a faithfulness gate in the SimpleMem~\cite{simplemem2026} long-term memory system.
\end{itemize}

\section{\framework{}: A Unified Toolkit}
\label{sec:toolkit}

\begin{figure*}[t]
\centering
\includegraphics[width=\linewidth]{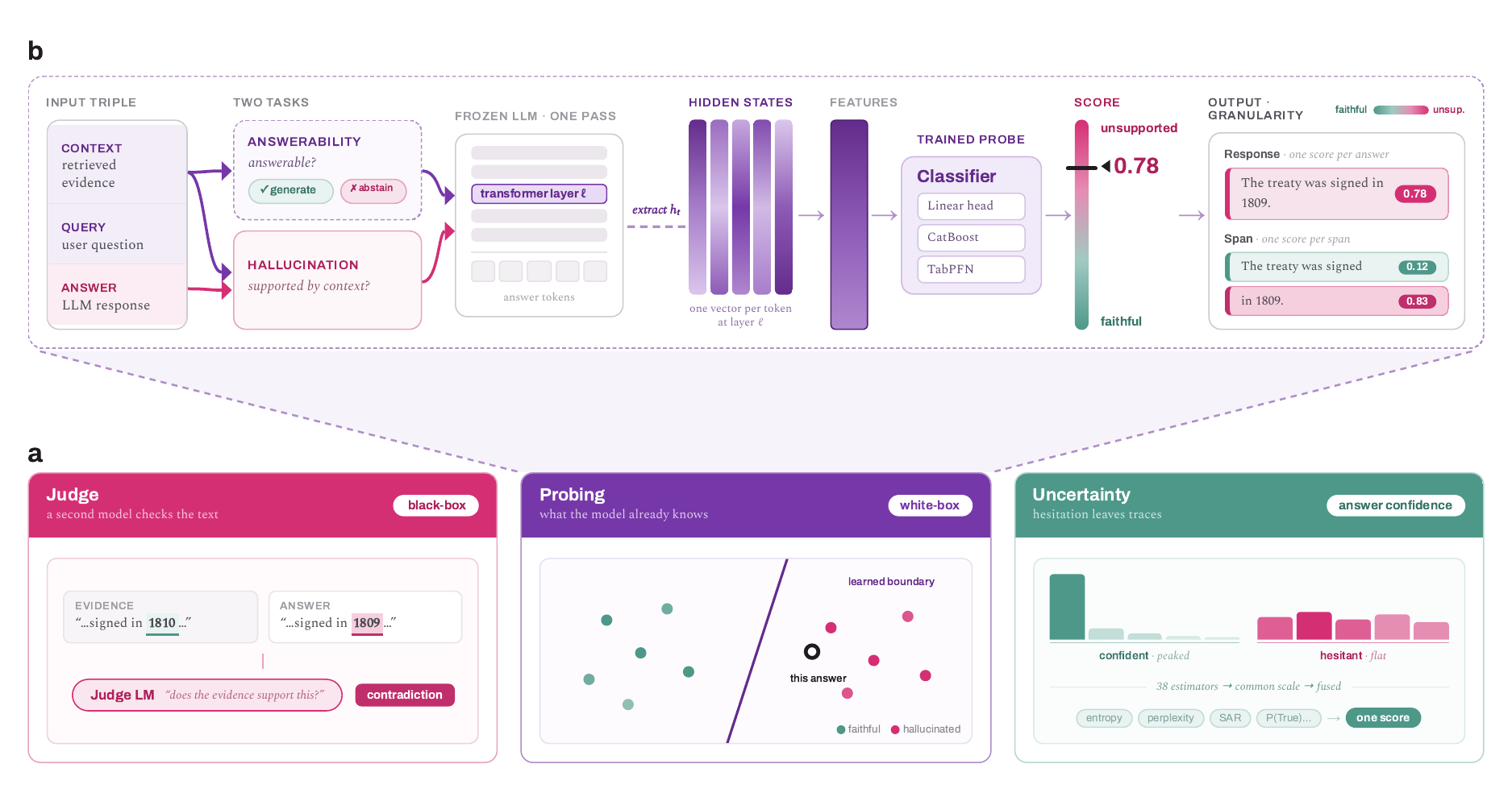}
\caption{\textbf{(a)} Three detector families: \emph{Judge} (black-box) uses a second model to check for contradictions between evidence and answer; \emph{Probing} (white-box) trains a classifier over hidden-state representations to separate faithful from hallucinated outputs; \emph{Uncertainty} fuses 38 estimators (entropy, perplexity, SAR, P(True), etc.) into a single calibrated score based on generation confidence patterns. \textbf{(b)} The full pipeline on \emph{Probing} detector example: an input triple context--query--answer is processed through a frozen LLM in one pass to extract hidden states, which are transformed into features and scored by a trained probe. Outputs can be at the response level (one score per answer) or span level (one score per text segment).}
\label{fig:arch}
\end{figure*}

\framework{} implements faithfulness inspection as a configurable pipeline over grounded inputs; Figure~\ref{fig:arch} gives an overview. Unlike toolkits focused only on post-generation hallucination detection, \framework{} supports two complementary tasks: \emph{contextual hallucination detection}, which verifies whether a generated answer is supported by the provided context, and \emph{query answerability}, which checks whether the context is sufficient before generation. Two tasks share the same interface but differ in their input:
\[
x_{\mathrm{hall}}=(c,q,a), \qquad x_{\mathrm{ans}}=(c,q),
\]
where \(c\) is the context, \(q\) is the user query, and \(a\) is the model answer. A detector \(D_{\theta}\) maps either input to a standardized inspection report
\[
D_{\theta}(x)=\big(s,\hat{y},E\big),
\]
where \(s\in[0,1]\) is a calibrated probability score, \(\hat{y}\) is the predicted label (binary, or ternary in NLI-style setups), and \(E\) contains optional evidence such as span scores, uncertainty values, or judge rationales. This shared report schema is the core abstraction of \framework{}: detectors and their ensembles can be compared in the Web UI or used as runtime gates without changing application code.

\paragraph{Detector families.}
\framework{} groups detectors into three families (Figure~\ref{fig:arch}a). \textbf{Probing detectors} train lightweight classifiers (linear, tabular, or attention-pooling, with optional contrastive objectives) over frozen internal features~\cite{hewitt2019control,ch-wang-etal-2024-androids, kossen2024semantic, belikova2025data,ji2024internal,chen2020simclr}. \textbf{Judge detectors} perform text-only semantic verification using encoder, decoder, or API-based models, covering NLI-style and alignment-style factuality checks~\cite{lettucedetect2025,ragtruth,zha2023alignscore}. \textbf{Uncertainty detectors} wrap \textsc{LM-Polygraph}\footnote{\url{https://github.com/IINemo/lm-polygraph}}, exposing likelihood-, sampling-, attention-based, and other uncertainty estimators. The families differ in the signal they consume and \framework{} makes these access assumptions explicit and allows implementation components to be interchanged.

\begin{figure*}[h!]
\centering
 \includegraphics[width=0.85\linewidth]{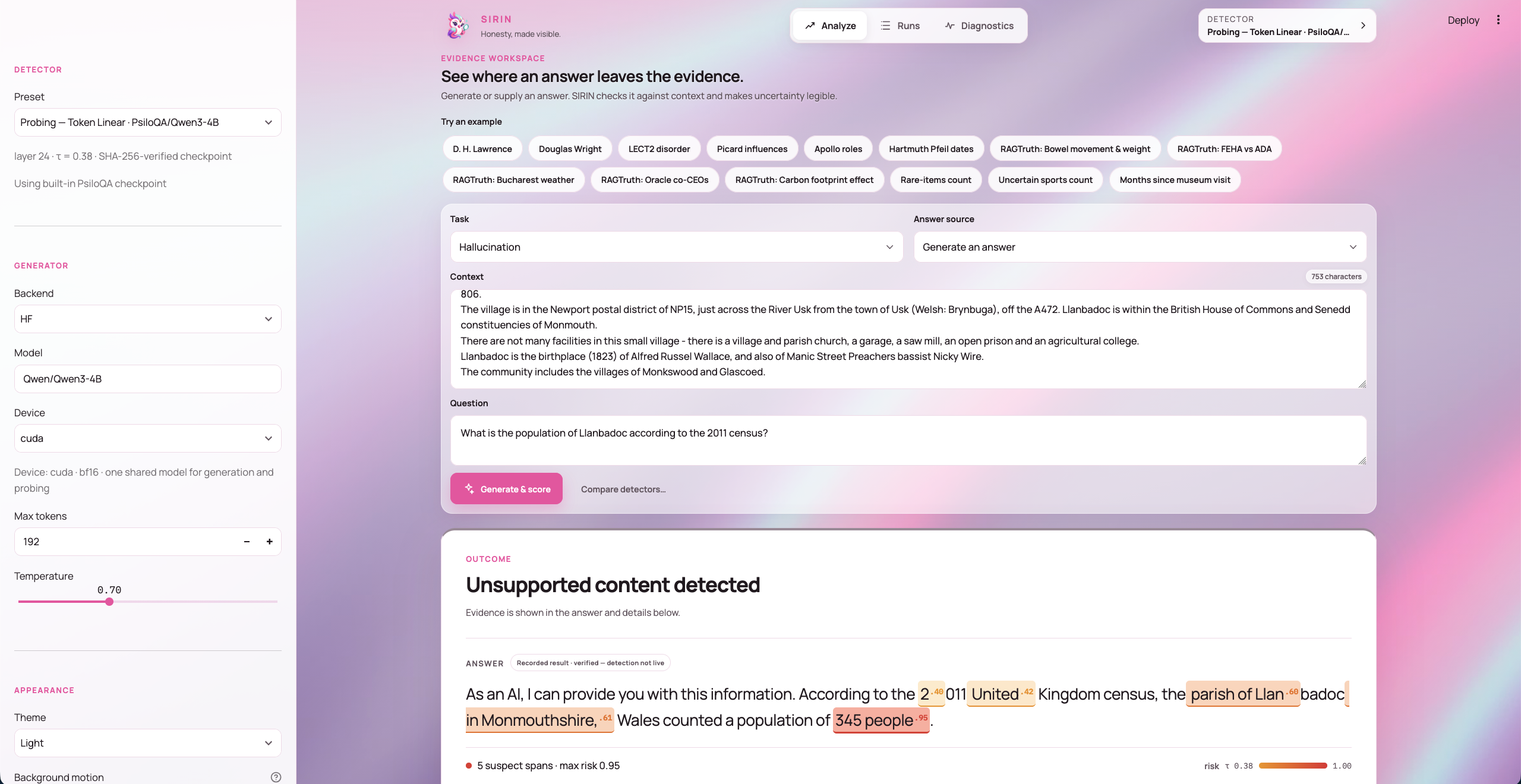}
\caption{\framework{}'s interactive Web UI. A visitor supplies a context--question pair
(or picks a curated example), selects a detector preset spanning the probing, judge,
and uncertainty families, and inspects the results directly: the answer is scored and
its unsupported spans are highlighted with per-span risk, alongside a summary of
suspect spans and maximum risk.}
\label{fig:ui-main}
\end{figure*}

\paragraph{Anatomy of a detection pass.}
Figure~\ref{fig:arch}b illustrates the probing pipeline: the input triple is rendered as a chat-formatted sample and processed by a frozen backend LLM in a \emph{single} forward pass. A feature processor extracts the signals; a token locator selects which positions to read and pools them into fixed-size features; a trained probe then emits a calibrated score. A composing processor extracts all requested feature families from the same shared pass, so multi-signal probes cost close to one model pass instead of one pass per signal. The same pipeline emits response-level scores for runtime gating or span-level scores for inspection via a reusable splitter/aggregator, following atomic decomposition~\cite{min2023factscore}, so one configuration serves both production gating and UI span highlighting.

\paragraph{Backends, configuration, and extension.}
\label{sec:ext}
Model calls are routed through interchangeable backends: \texttt{Transformers}\footnote{\url{https://github.com/huggingface/transformers}} for white-box access to internals, \texttt{vLLM}\footnote{\url{https://docs.vllm.ai/}} for high-throughput generation and log-probability scoring, and OpenAI-compatible endpoints for API-based black-box judges. Since each detector declares the signals it requires, the same pipeline runs with generator internals, with a proxy model when internals are unavailable, or as text-only verification through an API. A \framework{} run is a typed, Hydra\footnote{\url{https://hydra.cc/}}-compatible configuration over backend, processors, detector, splitter, and aggregation: switching task, detector family, or granularity is a configuration edit rather than a pipeline rewrite. New methods enter through three plug points: a feature processor, a detector, or an aggregator. Appendix~\ref{app:details} gives full implementation details.

\paragraph{Interactive Web UI.}
\label{sec:ui}
The UI is implemented directly in Streamlit\footnote{\url{https://streamlit.io}} and calls \framework{} detectors within the same process. Hugging Face and OpenAI-compatible backends provide generation. \framework{}'s thread-safe model manager reuses loaded models and evicts them when the configured model-count or GPU-memory limits are reached. Recorded JSON artifacts provide a GPU-free replay path.

A visitor pastes a context--query--answer triple or picks one from a curated gallery drawn from RAGTruth~\cite{ragtruth}, PsiloQA~\cite{psiloqa2025}, and Mu-SHROOM~\cite{mushroom2025}, then selects one of preloaded detector configurations spanning all three families and inspects the response-level score, span-level highlighting, per-chunk context heat-bar, and side-by-side compare mode directly in the interface (Figure~\ref{fig:ui-main}).


\section{Evaluation}
\label{sec:eval}

We evaluate \framework{} on response-level hallucination detection, query answerability, and span-level hallucination localization. The experiments compare probing, judge, and uncertainty-estimation methods under different model-access and annotation settings.

\paragraph{Setup.}
All detectors use the same score-and-calibration interface and differ only in the signals they consume. Probing methods train lightweight classifiers over frozen hidden states; encoder judges based on ModernBERT and DeBERTa-v3 are fine-tuned with LoRA, and the Qwen3-4B decoder judge is evaluated both zero-shot and after LoRA adaptation; GPT-5.4-mini is used as a strictly black-box API judge. Decision thresholds are selected on the training split to maximize F1; span metrics are computed at the character level. Appendix~\ref{app:setup} details the probe features, judge variants, and scoring protocol.

\subsection{Response-Level Detection}

\begin{table}[t]
\centering
\scriptsize
\setlength{\tabcolsep}{3.8pt}
\renewcommand{\arraystretch}{1.08}
\begin{tabular}{lcccccc}
\toprule
\multirow{2}{*}{\textbf{Method}}
& \multicolumn{2}{c}{\textbf{QA}}
& \multicolumn{2}{c}{\textbf{Summary}}
& \multicolumn{2}{c}{\textbf{Data-to-Text}} \\
\cmidrule(lr){2-3}\cmidrule(lr){4-5}\cmidrule(lr){6-7}
& AUROC & F1
& AUROC & F1
& AUROC & F1 \\
\midrule

\multicolumn{7}{l}{\textit{Probing}} \\
Att.-pool probe
& 86.9 & \second{60.6}
& \second{82.8} & \third{57.2}
& \third{85.3} & 83.6 \\
CatBoost probe
& 87.9 & 56.2
& 76.5 & 48.5
& \third{85.3} & \third{85.0} \\
CatBoost ensemble
& \third{88.0} & 53.0
& 77.7 & 52.1
& 84.9 & 83.3 \\

\midrule

\multicolumn{7}{l}{\textit{Judges}} \\
ModernBERT-large
& 80.2 & 48.9
& 62.2 & 27.5
& 79.6 & 82.6 \\
DeBERTa-v3-large
& 81.6 & 51.4
& 63.0 & 14.0
& 49.0 & 78.3 \\
Qwen3-4B (ZS)
& 76.1 & 43.9
& 64.2 & 41.1
& 79.0 & 75.1 \\
Qwen3-4B
& \best{93.5} & \best{71.4}
& \best{91.8} & \best{71.6}
& \best{93.4} & \best{91.0} \\
GPT-5.4-mini (ZS)
& \second{88.8} & \third{60.0}
& \third{82.0} & \second{57.3}
& \second{86.8} & \second{86.2} \\

\bottomrule
\end{tabular}
\vspace{-3pt}
\caption{Response-level contextual hallucination detection across the three RAGTruth task splits. Probing models operate on internal representations of Qwen3-4B.}
\label{tab:ragtruth}
\end{table}

\paragraph{RAGTruth.}
Table~\ref{tab:ragtruth} reports hallucination detection results on the question-answering (QA), summarization, and data-to-text subsets of RAGTruth~\cite{ragtruth}, using the original generated responses and expert annotations. Since the internal states of the original generators are unavailable, probing methods extract representations from Qwen3-4B over the annotated samples. Consistent with the annotation efficiency of lightweight probes~\cite{belikova2025data}, the best probe attains 88.0 AUROC on QA and 85.3 on data-to-text, outperforming the LoRA fine-tuned encoder judges on all three subsets (80.2--81.6 AUROC on QA); on summarization, where the encoders drop to 62.2--63.0 AUROC, the attention-pooling probe retains 82.8. Given sufficient annotations, however, the LoRA-trained Qwen3-4B judge is uniformly strongest (91.8--93.5 AUROC, 71.4--91.0 F1), 5.5--9.0 AUROC points above the best probe per subset, followed by zero-shot GPT-5.4-mini (82.0--88.8 AUROC). Uncertainty estimation is not reported because uncertainty signals from a separate proxy model are poorly aligned with the uncertainty of the original generator.

\begin{table}[h]
\centering
\scriptsize
\setlength{\tabcolsep}{4.8pt}
\renewcommand{\arraystretch}{1.08}
\begin{tabular}{@{}lcccc@{}}
\toprule
\multirow{2}{*}{\textbf{Method}}
& \multicolumn{2}{c}{\textbf{Hallucination Detection}}
& \multicolumn{2}{c}{\textbf{Question Answerability}} \\
\cmidrule(lr){2-3}\cmidrule(l){4-5}
& AUROC & F1 & AUROC & F1 \\
\midrule

\multicolumn{5}{l}{\textit{Probing}} \\
Att.-pool probe
& 75.3 & 46.0
& 87.9 & 79.5 \\
CatBoost probe
& \third{78.4} & 52.0
& 80.2 & 72.7 \\
CatBoost ensemble
& 74.0 & 46.7
& 78.6 & 71.7 \\

\midrule

\multicolumn{5}{l}{\textit{Judges}} \\
ModernBERT-large
& 73.7 & 32.5
& 50.0 & 49.6 \\
DeBERTa-v3-large
& 77.7 & \third{52.4}
& \third{91.3} & \second{84.6} \\
Qwen3-4B (ZS)
& 63.6 & 28.6
& \second{92.4} & 78.2 \\
Qwen3-4B
& \best{87.0} & \best{70.2}
& \best{95.7} & \best{89.2} \\
GPT-5.4-mini (ZS)
& \second{84.6} & \second{61.0}
& \third{91.3} & \third{83.8} \\

\midrule

\multicolumn{5}{l}{\textit{Uncertainty Estimation}} \\
MeanTokenEntropy
& 67.4 & 39.9
& 52.0 & 57.8 \\
RAUQ
& 58.8 & 39.8
& 58.8 & 51.2 \\
Focus
& 64.0 & 39.2
& 56.9 & 52.8 \\
Combined UE
& 69.0 & 39.9
& 57.8 & 58.6 \\

\bottomrule
\end{tabular}
\vspace{-3pt}
\caption{Hallucination detection and question answerability on SQuAD~2.0. Probing and uncertainty estimation use internal signals from Qwen3-4B.}
\label{tab:nonproxy}
\end{table}

\paragraph{SQuAD~2.0.}
Table~\ref{tab:nonproxy} evaluates hallucination detection using responses generated by Qwen3-4B for a SQuAD~2.0 subset~\cite{rajpurkar2018know}. This allows probing and uncertainty methods to access signals from the same model that produced the answer. Probes provide a lightweight supervised baseline of up to 78.4 AUROC, but the LoRA-trained Qwen3-4B judge again performs best in this annotation-rich setting (87.0 AUROC, 70.2 F1), with zero-shot GPT-5.4-mini close behind (84.6 AUROC). The uncertainty methods are weaker but fully unsupervised: the three top-performing estimators, MeanTokenEntropy, RAUQ, and Focus~\cite{fomicheva-etal-2020-unsupervised,vazhentsevefficient,zhang-etal-2023focus}, reach 58.8--67.4 AUROC, and a fused score (Combined UE), which normalizes and aggregates the best-performing estimators on the training split, adds only 1.6 points (69.0 AUROC).

We additionally evaluate \emph{query answerability} using the original SQuAD~2.0 labels and only the question and context as input. The generic ModernBERT encoder fails to transfer, scoring at chance level (50.0 AUROC), whereas DeBERTa-v3 with NLI pretraining reaches 91.3 AUROC, within 4.4 points of the LoRA-trained Qwen3-4B judge (95.7) and on par with zero-shot GPT-5.4-mini (91.3). This suggests that answerability is closely aligned with textual entailment. The attention-pooling probe attains 87.9 AUROC, demonstrating that the same \framework{} pipeline can support both pre-generation answerability checks and post-generation faithfulness verification.

\subsection{Span-Level Detection}

\begin{table}[h!]
\centering
\small
\setlength{\tabcolsep}{5pt}
\renewcommand{\arraystretch}{1.08}
\begin{tabular}{lcccc}
\toprule
\textbf{Method} & AUROC & AP & F1 & IoU \\
\midrule

\multicolumn{5}{l}{\textit{Probing}} \\
Att.-pool probe & \best{76.0} & \third{65.6} & 72.5 & 59.6 \\
CatBoost probe          & \third{75.9} & 65.4 & \third{73.2} & 61.2 \\
CatBoost ensemble       & \third{75.9} & \second{65.8} & 72.2 & 60.3 \\

\midrule

\multicolumn{5}{l}{\textit{Judges}} \\
ModernBERT-large & 74.6 & 64.5 & \second{73.4} & \third{63.5} \\
DeBERTa-v3-large & 74.4 & 64.8 & \third{73.2} & \third{63.2} \\
Qwen3-4B (ZS) & 50.0 & 51.3 & 0.0 & 3.7 \\
Qwen3-4B & \second{75.8} & \best{68.4} & \best{74.8} & \best{69.3} \\
GPT-5.4-mini (ZS) & 63.8 & 58.5 & 72.4 & 59.6 \\
\bottomrule
\end{tabular}
\vspace{-3pt}
\caption{Span-level hallucination detection on the English subset of PsiloQA. Probing models use Qwen3-4B internal representations; all metrics are character-level, including AUROC, AP, F1, and mean per-sample IoU.}
\label{tab:spanlevel}
\end{table}

\begin{table*}[htb]
\centering
\small
\setlength{\tabcolsep}{4pt}
\renewcommand{\arraystretch}{1.0}
\begin{tabular}{ll cccc cc r}
\toprule
\multirow{2}{*}{\textbf{Detector}}
& \multirow{2}{*}{\textbf{Condition}}
& \multicolumn{4}{c}{\textbf{Served accuracy (\%)}}
& \multirow{2}{*}{\makecell{\textbf{Strict}\\\textbf{unfaith. (\%)}}}
& \multirow{2}{*}{\makecell{\textbf{Abstain}\\\textbf{rate (\%)}}}
& \multirow{2}{*}{\makecell{$\Delta$\textbf{tokens}\\\textbf{LLM gen.}}} \\
\cmidrule(lr){3-6}
& & Single & Multi & Temporal & Overall & & & \\
\midrule

-- & Memory baseline
& 70.1 & 58.7 & 49.6 & 62.4 & 20.2 & 0.0 & 0\% \\

Oracle & Correctness @80\% cov.
& -- & -- & -- & 78.0 & -- & 20.0 & 0\% \\

\midrule

\multirow{3}{*}{\shortstack[l]{\textbf{SIRIN Probing}}}
& $+$ Answerability gate
& 78.4
& 60.0
& 63.2
& 69.9
& 17.3
& 21.2
& $-$5.6\% \\

& $+$ Faithfulness gate
& 78.7
& \third{64.4}
& 61.7
& 71.7
& 12.9
& 20.0
& 0\% \\

& $+$ Both gates
& \best{85.0}
& 65.8
& \best{70.4}
& \second{76.8}
& \second{11.6}
& 33.0
& $-$5.6\% \\

\midrule

\multirow{3}{*}{\shortstack[l]{\textbf{SIRIN Judge}}}
& $+$ Answerability gate
& \third{82.5}
& 62.3
& \third{64.8}
& \third{72.5}
& 17.6
& 21.2
& $-$86.2\% \\

& $+$ Faithfulness gate
& 71.2
& \second{71.7}
& 53.4
& 68.8
& \third{12.1}
& 20.0
& 0\% \\

& $+$ Both gates
& \second{84.5}
& \best{75.3}
& \second{69.8}
& \best{79.3}
& \best{9.8}
& 35.9
& $-$86.2\% \\

\bottomrule
\end{tabular}
\caption{\framework{} as a selective-prediction layer for SimpleMem on LongMemEval. Qwen3.5-35B-A3B generates all answers. \framework{} uses either probes over its hidden states or GPT-5.4-mini as a text-only judge. $\Delta$tokens is relative to the ungated baseline, includes generator model completion tokens, and excludes detector inference.}
\label{tab:memory}
\end{table*}

Table~\ref{tab:spanlevel} reports character-level hallucination localization on the English subset of PsiloQA~\cite{psiloqa2025}. As the original generator internals are unavailable, probing methods use Qwen3-4B representations extracted from the annotated samples. Probes lead in ranking quality (up to 76.0 AUROC vs.\ 75.8 for the best judge) and remain attractive when span annotations are scarce. With richer supervision, however, the LoRA-trained Qwen3-4B judge achieves the strongest localization (74.8 F1, 69.3 IoU, vs.\ at most 73.2 F1 and 61.2 IoU among probes), while zero-shot span tagging collapses entirely. These results indicate that proxy probing is useful for annotation-efficient localization, whereas trained judges are preferable when a sufficiently large labeled dataset is available.

\section{Trustworthy memory with \framework{}}
\label{sec:simplemem}

\begin{figure}[h!]
\centering
\includegraphics[width=\columnwidth]{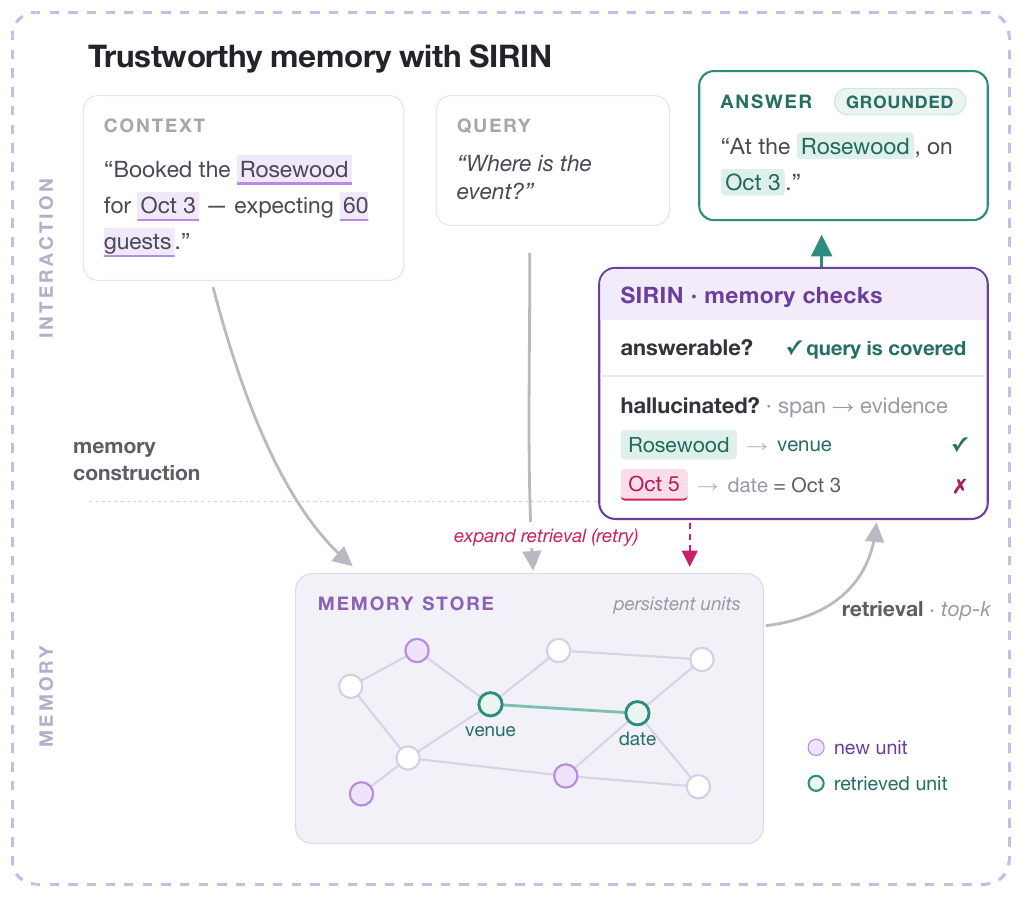}
\caption{\framework{} for trustworthy dynamic memory. Answerability and span-level faithfulness checks validate retrieved evidence before returning an answer or trigger retrieval expansion on failure.}
\label{fig:simplemem}
\end{figure}

Long-term memory turns hallucination from an isolated mistake into a compounding one: an agent that misreads its own memory confidently contradicts facts the user trusts it to remember. Since every agentic memory framework ultimately serves a retrieval stage that emits evidence for answering a query, \framework{} can attach at that stage as a drop-in trust layer, with no changes to the memory system itself.

Figure~\ref{fig:simplemem} shows the resulting loop: incoming messages are distilled into memory units, and at question time \framework{} applies two gates around retrieval. The \emph{answerability} gate (pre-generation) checks whether retrieved units cover the query, expanding retrieval on failure; the \emph{hallucination} gate (post-generation) validates each answer span against retrieved evidence and triggers a retry on contradiction 
Together the gates give the agent three actions (\emph{serve}, \emph{retry}, or \emph{abstain}) turning hallucination detection into selective prediction.

We instantiate this design in SimpleMem~\cite{simplemem2026} and evaluate it on LongMemEval using lightweight attention-pooling probes over the generator's hidden states and GPT-5.4-mini as a text-only judge (Table~\ref{tab:memory}). Combining answerability and faithfulness checks raises accuracy on served answers from the 62.4\% ungated baseline to 79.3\% while reducing strict unfaithfulness from 20.2\% to 9.8\%. This demonstrates that \framework{} can provide an effective trust layer for memory-grounded agents. Appendices~\ref{app:memory_frameworks} and~\ref{app:probe_auroc} extend this to memory systems Mem0~\cite{mem0} and LightMem~\cite{lightmem}.

\section{Conclusion}
Contextual hallucination detection is an infrastructure problem, not a single-detector problem, and \framework{} treats it that way: probing, judge-style, and uncertainty-estimation detectors, together with pre-generation query answerability, share one interface, one report schema, and a web UI where they can be inspected and compared live. The payoff is concrete: lightweight probes recover most of the detection signal when labels are scarce, trained judges dominate when supervision is plentiful, and two selective gates lift a memory-grounded agent from 62.4\% to 79.3\% served accuracy while halving flagged unfaithfulness, at no extra generation cost. We see \framework{} as a step toward verification becoming a standard layer of the LLM stack: every grounded response checked, every unsupported span visible, and every new detector one plug-in away. \framework{} is released under Apache~2.0 at \url{https://github.com/sb-ai-lab/SIRIN}, with the Web UI, configuration presets, and memory-integration examples.




\section*{Limitations}
\label{sec:limitations}
\framework{}'s probing and judge detectors are trained on English data and have not been validated in multilingual settings. Detector accuracy still depends on the quality and domain match of the annotated data used to fit probes and fine-tune judges, so out-of-domain deployment may require re-calibration. The memory gating results are evaluated with a single generator backbone, using probes over its hidden states and one API judge as gates; we have not yet evaluated gate transfer across unrelated backbones without refitting. Finally, the interactive Web UI currently keeps a fixed number of model configurations warm in memory (6--8 on one A100), which bounds how many detectors can be compared live without additional hardware.

\bibliography{custom}

\appendix
\section{Toolkit Implementation Details}
\label{app:details}

This appendix expands on the implementation mechanisms summarized in \S\ref{sec:toolkit}.

\paragraph{Feature processors.}
A feature processor maps a batch of chat-formatted samples to tensors. \framework{} supports hidden states, attention maps, logits, sublayer outputs, Lookback Lens-style attention ratios, topological features over attention graphs, and token- or sequence-level uncertainty scores~\cite{lookback2024,bazarova2026hallucination,fadeeva2023}. Formally, a processor computes
\[
\phi_k(x;M,\ell)\rightarrow z_k,
\]
where \(M\) is the selected backend model, \(\ell\) is an optional layer index, and \(z_k\) is a feature block. A composing processor aligns layers, padding, and masks across feature families and extracts compatible blocks from one shared forward pass, so multi-signal probes cost close to one model pass instead of one pass per signal.

\paragraph{Token location and pooling.}
White-box detectors often depend on where internal states are read: the last generated token, the end-of-sequence token, a response boundary, or a window around the answer. \framework{} separates \emph{where to look} from \emph{what to extract}. A token locator resolves answer-token positions, boundary offsets, substring anchors, and local neighborhoods; a pooling operator then maps selected token states to fixed-size features:
\[
z=\operatorname{pool}\big(\phi(x)_{\mathcal{T}}\big), \qquad
\mathcal{T}=\operatorname{locate}(x).
\]
This makes method-specific assumptions explicit in configuration. For example, a Lookback Lens-style detector is specified by choosing the context--answer boundary and the attention-ratio processor~\cite{lookback2024}, while attention-pooling probes can learn how to aggregate distributed token or layer evidence~\cite{lin2017selfattentive,ilse2018attentionmil}.

\paragraph{Splitting and aggregation.}
A splitter decomposes the context, the response, or both into character windows, sentences, paragraphs, token-aware chunks, \texttt{LangChain}\footnote{\url{https://python.langchain.com/}} chunks, or LLM-extracted atomic claims. Any detector can score the resulting segments. Segment scores \(\{s_i\}_{i=1}^{n}\) are folded back into a response-level score by
\[
s=\operatorname{aggregate}(s_1,\ldots,s_n),
\]
where \(\operatorname{aggregate}\) can be mean, maximum, minimum, majority vote, fixed weighting, or learned fusion.

\paragraph{Efficiency and reproducibility.}
\framework{} exposes efficiency mechanisms as interface features. Local backends process large datasets in fixed-size chunks, validate returned tensors, and merge partial outputs. API backends use bounded asynchronous fan-out with retry logic. Extracted features are cached by canonicalized sample digest, model identifier, layer, and feature type, so adding a layer, enabling a new feature family, or resuming an interrupted run recomputes only missing artifacts. Probe features are normalized per block and adaptively compressed with PCA or \texttt{UMAP}\footnote{\url{https://umap-learn.readthedocs.io/}} only when dimensionality exceeds a configured threshold. These details keep detector sweeps practical and let UI presets stay warm, while inference reuses the same serialized transforms as training.

\paragraph{Logging.}
Logging is pluggable through Weights \& Biases\footnote{\url{https://wandb.ai/}}, ClearML\footnote{\url{https://clear.ml/}}, and other simultaneous trackers; the full run configuration (backend, processors, token locator, detector, splitter, calibration rule, and aggregation strategy) is serialized with every run.

\section{Full Feature Table}
\label{app:features}

Table~\ref{tab:features} summarizes the main interface features of \framework{}.

\begin{table}[ht]
\centering
\begin{tcolorbox}[
  enhanced,
  arc=3pt,
  boxrule=0.45pt,
  colback=white,
  colframe=sirinrule,
  coltext=sirintext,
  left=5pt,
  right=5pt,
  top=4pt,
  bottom=3pt,
  width=\columnwidth
]
\footnotesize
\renewcommand{\arraystretch}{1.12}
\setlength{\tabcolsep}{4pt}
\arrayrulecolor{sirinrule}

\begin{tabular}{@{}>{\raggedright\arraybackslash\itshape}p{0.31\columnwidth}
                  >{\raggedright\arraybackslash}p{0.61\columnwidth}@{}}
\rowcolor{sirinhl}
{\normalfont\bfseries Highlight} & \textbf{Why does it matter?} \\
\specialrule{0.45pt}{0pt}{2pt}

One report schema
& Probes, judges, uncertainty estimators, and ensembles all produce the same calibrated inspection object. \\

\addlinespace[2pt]
White-box \(\leftrightarrow\) black-box
& \texttt{Transformers}, \texttt{vLLM}, and API backends make access assumptions explicit and swappable. \\

\addlinespace[2pt]
One-pass multi-signal probes
& Hidden states, logits, attention, lookbacks, topology, and UE scores can be composed without repeated extraction. \\

\addlinespace[2pt]
Configurable token location
& Last-token probes, boundary probes, substring anchors, and Lookback Lens-style ratios become config choices. \\

\addlinespace[2pt]
Any detector becomes local
& Sentence, paragraph, token-aware, and atomic-claim splitting enables span inspection from response-level methods. \\

\addlinespace[2pt]
UE catalog in one API
& \textsc{LM-Polygraph} estimators, calibration, and fusion are exposed as regular \framework{} detectors. \\

\addlinespace[2pt]
Built for sweeps
& Batching, feature caching, adaptive compression, resumability, and logging make ablations and UI presets practical. \\

\end{tabular}
\arrayrulecolor{black}
\end{tcolorbox}
\vspace{-4pt}
\caption{\framework{} interface highlights. The toolkit emphasizes reusable detector abstractions rather than a single hallucination-detection method.}
\label{tab:features}
\end{table}




\section{Experimental Setup Details}
\label{app:setup}

Probing methods are trained on the annotated split over hidden states taken from the middle layer and attention-pooled across tokens~\cite{ch-wang-etal-2024-androids}; they comprise an $\ell_2$-regularized logistic regression, CatBoost classifiers over hidden states mean-pooled across tokens, and an ensemble that additionally incorporates lookback-ratios features~\cite{lookback2024}. Encoder judges based on ModernBERT\footnote{\url{https://hf.co/answerdotai/ModernBERT-large}} and DeBERTa-v3\footnote{\url{https://hf.co/microsoft/deberta-v3-large}} are fine-tuned with LoRA, and the Qwen3-4B\footnote{\url{https://hf.co/Qwen/Qwen3-4B}} decoder judge is evaluated both zero-shot and after LoRA adaptation. GPT-5.4-mini receives only the textual input and is scored from the probability it assigns to the positive label. For span localization, decoder and API judges emit explicit span tags.

\section{Detector Discrimination Across Memory Frameworks}
\label{app:probe_auroc}

Appendix~\ref{app:memory_frameworks} evaluates the gates end-to-end, whereas this section measures how well the probe separates the labeled construct, independent of how a gate uses the resulting ranking. We hold the probe configuration and the relabeling fixed across all three memory systems, so each task's three rows are directly comparable.

\begin{table}[ht]
    \centering
    \small
    \setlength{\tabcolsep}{4pt}
    \resizebox{\columnwidth}{!}{%
    \begin{tabular}{ll ccc}
    \toprule
    \textbf{Task} & \textbf{Memory system} & \textbf{AUROC} & \textbf{AP} & \textbf{Prev.} \\
    \midrule
    \multirow{3}{*}{Answerability}
    & SimpleMem & 0.843 \small{[.81,.88]} & 0.762 & 0.356 \\
    & Mem0      & \textbf{0.913} \small{[.89,.94]} & 0.935 & 0.566 \\
    & LightMem  & 0.884 \small{[.85,.91]} & 0.860 & 0.407 \\
    \midrule
    \multirow{3}{*}{Faithfulness}
    & SimpleMem & 0.793 \small{[.74,.84]} & 0.641 & 0.226 \\
    & Mem0      & \textbf{0.876} \small{[.84,.91]} & 0.692 & 0.262 \\
    & LightMem  & 0.844 \small{[.81,.88]} & 0.655 & 0.254 \\
    \bottomrule
    \end{tabular}%
    }
    \caption{Probe discrimination for the three memory systems on the shared appendix relabelling, one probe configuration throughout. Layer~30 is selected in 27 of 30 folds. Intervals are 95\% dependency-cluster bootstrap. AUROC is the headline because it is invariant to prevalence, which ranges from $0.226$ to $0.566$ here. AP is reported beside its prevalence and should \emph{not} be compared across systems for that reason. Restricting to the cohort labeled in all three systems ($n{=}474$ for answerability, $500$ for faithfulness) leaves every value within $0.01$ and the ordering unchanged.}
    \label{tab:probe_auroc}
    \end{table}

The probe discriminates well on every system and \emph{best} on Mem0. Together with Table~\ref{tab:memory_appendix}, these results separate two easily conflated quantities: a detector's accuracy about its label and the served accuracy recovered by a gate. Detector AUROC alone therefore does not predict the value of a gate. The coverage headroom and label--correctness bounds in Appendix~\ref{app:memory_frameworks} cap what any gate can recover.

\section{Trustworthy Memory on Additional Memory Frameworks}
\label{app:memory_frameworks}

Table~\ref{tab:memory} builds the selective-prediction layer on SimpleMem. To test whether it depends on that memory framework, we run the same two gates over Mem0~\cite{mem0} and LightMem~\cite{lightmem}, holding the generator, the probe configuration and the 80\% coverage target fixed. Only the memory framework changes.

\begin{table*}[htb]
\centering
\small
\setlength{\tabcolsep}{4pt}
\renewcommand{\arraystretch}{1.0}
\begin{tabular}{ll cccc cc r}
\toprule
\multirow{2}{*}{\textbf{Detector}}
& \multirow{2}{*}{\textbf{Condition}}
& \multicolumn{4}{c}{\textbf{Served accuracy (\%)}}
& \multirow{2}{*}{\makecell{\textbf{Strict}\\\textbf{unfaith. (\%)}}}
& \multirow{2}{*}{\makecell{\textbf{Abstain}\\\textbf{rate (\%)}}}
& \multirow{2}{*}{\makecell{$\Delta$\textbf{tokens}\\\textbf{LLM gen.}}} \\
\cmidrule(lr){3-6}
& & Single & Multi & Temporal & Overall & & & \\
\midrule
\multicolumn{9}{l}{\emph{(a) Mem0} ($N{=}500$)} \\
\midrule

-- & Memory baseline
& 41.0 & 31.6 & 26.3 & 37.8 & 35.1 & 0.0 & 0\% \\

Oracle & Correctness @80\% cov.
& -- & -- & -- & 47.2 & -- & 20.0 & 0\% \\

\midrule

\multirow{3}{*}{\shortstack[l]{\textbf{SIRIN Probing}}}
& $+$ Answerability gate
& \third{49.2} & 34.9 & 28.9 & \third{43.5} & 36.1 & 20.0 & $-$17.0\% \\

& $+$ Faithfulness gate
& 40.6 & 39.1 & 23.2 & 39.8 & \second{25.3} & 20.0 & 0\% \\

& $+$ Both gates
& 48.8 & \second{43.1} & \third{29.4} & \second{46.9} & \third{27.1} & 36.4 & $-$17.0\% \\

\midrule

\multirow{3}{*}{\shortstack[l]{\textbf{SIRIN Judge}}}
& $+$ Answerability gate
& \second{52.6} & 33.0 & \second{31.1} & \third{43.5} & 40.9 & 20.0 & $-$19.8\% \\

& $+$ Faithfulness gate
& 41.4 & \third{42.1} & 27.8 & 42.8 & \best{23.2} & 20.0 & 0\% \\

& $+$ Both gates
& \best{53.6} & \best{45.9} & \best{35.8} & \best{51.0} & 28.7 & 38.0 & $-$19.8\% \\

\midrule
\multicolumn{9}{l}{\emph{(b) LightMem} ($N{=}500$)} \\
\midrule

-- & Memory baseline
& 48.1 & 63.9 & 31.6 & 50.2 & 33.1 & 0.0 & 0\% \\

Oracle & Correctness @80\% cov.
& -- & -- & -- & 62.7 & -- & 20.0 & 0\% \\

\midrule

\multirow{3}{*}{\shortstack[l]{\textbf{SIRIN Probing}}}
& $+$ Answerability gate
& 61.6 & 64.4 & 33.0 & 54.2 & 32.9 & 20.0 & $-$23.3\% \\

& $+$ Faithfulness gate
& 49.0 & \third{70.5} & 31.1 & 50.0 & \second{26.1} & 20.0 & 0\% \\

& $+$ Both gates
& \best{63.2} & 69.8 & 34.5 & 55.3 & \third{26.3} & 36.4 & $-$23.3\% \\

\midrule

\multirow{3}{*}{\shortstack[l]{\textbf{SIRIN Judge}}}
& $+$ Answerability gate
& \third{62.7} & 64.4 & \third{36.4} & \second{56.5} & 32.6 & 20.0 & $-$26.0\% \\

& $+$ Faithfulness gate
& 49.3 & \second{78.6} & \second{37.6} & \third{56.2} & \best{22.1} & 20.0 & 0\% \\

& $+$ Both gates
& \second{63.0} & \best{79.3} & \best{48.5} & \best{64.5} & \best{22.1} & 36.4 & $-$26.0\% \\

\bottomrule
\end{tabular}
\caption{\framework{} as a selective-prediction layer for \textbf{Mem0}~\cite{mem0} and
\textbf{LightMem}~\cite{lightmem} on LongMemEval ($N{=}500$ each), in the format of
Table~\ref{tab:memory}. Qwen3.5-35B-A3B generates all answers. \framework{} uses either probes over the generator's hidden states or GPT-5.4-mini as a text-only judge. Single gates use a retrospective rank cut to target 80\% coverage. Appendix~\ref{app:memory_frameworks} lists how these panels differ from Table~\ref{tab:memory} and the caveats both tables share. Both gate intersections beat their ungated baseline under a dependency-cluster bootstrap (10{,}000 resamples). On Mem0, Probing $+$ Both gains $+9.1$ points (CI95 $[+5.9, +12.3]$) and Judge $+$ Both $+13.2$ points ($[+10.1, +16.4]$). On LightMem, the gains are $+5.1$ ($[+1.9, +8.4]$) and $+14.3$ ($[+11.0, +17.7]$). Unlike on SimpleMem, the judge significantly outperforms the probe on both systems (paired Probing $-$ Judge $\Delta$: Mem0 CI95 $[-7.8, -0.4]$; LightMem $[-13.1, -5.1]$), while the answerability gates alone are indistinguishable (Mem0 $[-2.1, +2.2]$; LightMem $[-4.5, +0.0]$). As in Table~\ref{tab:memory}, Overall includes all LongMemEval question categories and Knowledge-Update is not shown as a column.}
\label{tab:memory_appendix}
\end{table*}

These panels do not use the exact Table~\ref{tab:memory} pipeline. They apply a revised relabelling on the full $N{=}500$. Table~\ref{tab:memory} instead validates evidence quotes and drops 24 SimpleMem rows to $N{=}476$, which puts its answerability abstention at 21.2\% instead of 20.0\%. Its Strict-unfaith column is the probe's own training label, whereas here that column comes from an independent Claude Sonnet 5 judge. The coverage threshold, as in Table~\ref{tab:memory}, is a transductive within-evaluation rank cut.

\paragraph{Findings.}
The two gates carry over to both frameworks. Combined, they raise served accuracy by $+13.2$ points on Mem0 and $+14.3$ on LightMem under the judge, and the caption's bootstrap intervals exclude zero. The faithfulness gate also lowers served unfaithfulness. Because that column is scored independently of the probe, the reduction is real rather than an artifact of its own label: under the judge it falls from $35.1$ to $23.2$ on Mem0 and from $33.1$ to $22.1$ on LightMem. The probe transfers less well. Its faithfulness gate is flat on LightMem and adds only $+2.0$ points on Mem0, so most of its lift comes from the answerability gate. These panels are \emph{not} a ranking of memory frameworks: the ungated baselines differ too widely, so read each within its own system.

\end{document}